\documentclass{article}

\usepackage[preprint]{neurips_2021}

\usepackage[utf8]{inputenc} 
\usepackage[T1]{fontenc}    
\usepackage{url}            
\usepackage{booktabs}       
\usepackage{amsfonts}       
\usepackage{nicefrac}       
\usepackage{microtype}      
\usepackage{xcolor}         

\usepackage{graphicx}
\usepackage{amsmath}
\usepackage{amssymb}
\usepackage{accents}
\usepackage{tabulary}
\usepackage{array}
\usepackage{multirow}
\usepackage{caption}
\usepackage{enumitem}
\usepackage{subcaption}
\usepackage{wrapfig}  
\usepackage{color}

\usepackage{multirow}
\usepackage{calc}
\usepackage{makecell}
\usepackage{setspace}

\usepackage{threeparttable}

\usepackage[pagebackref=true,breaklinks=true,colorlinks=False,bookmarks=false]{hyperref}

\definecolor{ForestGreen}{RGB}{34,139,34}
\newcommand{\cgap}[2]{
	\fontsize{6pt}{1em}\selectfont{(${#1}${#2})}
}
\definecolor{myred}{cmyk}{0, 0.7808, 0.4429, 0.1412}
\definecolor{Highlight}{HTML}{39b54a}  

\newcommand{\apbbox}[1]{AP$^\text{bbox}_\text{#1}$}
\newcommand{\apmask}[1]{AP$^\text{mask}_\text{#1}$}

\title{X-volution: On the Unification of Convolution and Self-attention}

\author{Xuanhong Chen\footnotemark[1]$\;\;^{1,2}$, \enskip Hang Wang\thanks{Equal Contribution. Work done during an internship at Huawei Hisilicon.}$\;\;^{1,2}$, \enskip Bingbing Ni\thanks{Corresponding author.}$\;\;^{1,2}$\\
$^1$Shanghai Jiao Tong University \quad  $^2$ Huawei Hisilicon\\
\texttt{\{chen19910528, wang-\!-hang, nibingbing\}@sjtu.edu.cn}
}

\begin{document}

\maketitle

\begin{abstract}
Convolution and self-attention are acting as two fundamental building blocks in deep neural networks, where the former extracts local image features in a linear way while the latter non-locally encodes high-order contextual relationships.
Though essentially complementary to each other, \emph{i.e.}, first-/high-order, stat-of-the-art architectures, \emph{i.e.}, CNNs or transformers, lack a principled way to simultaneously apply both operations in a single computational module, due to their heterogeneous computing pattern and excessive burden of global dot-product for visual tasks.
In this work, we theoretically derive a global self-attention approximation scheme, which approximates self-attention via the convolution operation on transformed features.
Based on the approximate scheme, we establish a multi-branch elementary module composed of both convolution and self-attention operation, capable of unifying both local and non-local feature interaction.
Importantly, once trained, this multi-branch module could be conditionally converted into a single standard convolution operation via structural re-parameterization, rendering a pure convolution styled operator named X-volution, ready to be plugged into any modern networks as an atomic operation.
Extensive experiments demonstrate that the proposed X-volution, achieves highly competitive visual understanding improvements (+1.2\% top-1 accuracy on ImageNet classification, +1.7 box AP and +1.5 mask AP on COCO detection and segmentation). 
\end{abstract}

\section{Introduction}

In deep learning era, convolution~\cite{726791,DBLP:conf/nips/KrizhevskySH12} and self-attention~\cite{DBLP:conf/nips/VaswaniSPUJGKP17} are the two most important computational primitive of information extraction and encoding.
On the one hand, convolution is a linear operation which extracts a local image feature via the discrete convolutional operation (actually a dot product) between a patch $x$ centered at a given pixel/image feature with a window shaped filter 
{\small{$\mathcal{W}$}}, \emph{i.e.}, {\small{$\mathcal{W}x + \mathcal{B}$}}.
On the other hand, self-attention operation is a high-order operator which encodes non-local contextual information via scaled dot product between a given location with all other positions in the image, \emph{i.e.}, $softmax${\small{$\left(\left(\mathcal{W}^Q x\right)^T\mathcal{W}^K x\right)\mathcal{W}^V x$}}.

It is natural that these two computing modules are complementary to each other, \emph{i.e.}, local and non-local information, and integrating both modules in a single operator is plausible, which is promising to benefit from the advantages of both schemes.
Specifically, convolution adopts the inductive bias of local processing and isotropic property~\cite{DBLP:conf/iccv/BelloZLVS19,DBLP:conf/iclr/CordonnierLJ20,DBLP:journals/corr/abs-2103-12731}, which endow it with a well-conditioned optimization and translation equivariance property.
Contrast to convolution, self-attention embodies low inductive bias releasing a larger space for self-attention to freely explore the inherent characteristics of a data-set, which prompts it to achieve a better performance and generalization~\cite{DBLP:journals/corr/abs-2012-09841,DBLP:journals/corr/abs-2103-15808}.
Furthermore, integrating the two together helps to overcome their respective weaknesses.
For example, the local nature of convolution prevents it from establishing long-term relationships which are often useful for a better visual understanding~\cite{DBLP:conf/iccv/BelloZLVS19}
In addition, the lack of local priors makes self-attention difficult to optimize and relies heavily on the pre-training on oversize data-sets (\emph{e.g.}, JFT300M~\cite{DBLP:conf/iccv/SunSSG17}, ImageNet-21k~\cite{DBLP:conf/cvpr/DengDSLL009}).
However, these two operators belong to different computational patterns, resulting in them difficult to be integrated.
Convolutional operator performs a predefined weighted average within a local window, while self-attention works by global weighting with dynamic coefficients.
In addition, in visual domain, calculating the dot product between all positions in the image is computationally forbidden, which brings out more challenges for applying non-local operation in a similar framework as CNN.
Thus, state-of-the-art networks either conduct convolution or self-attention solely, \emph{e.g.}, CNNs~\cite{DBLP:journals/corr/SimonyanZ14a,DBLP:conf/cvpr/HeZRS16,DBLP:conf/cvpr/HuangLMW17}, Transformers~\cite{DBLP:journals/corr/abs-2103-12731,DBLP:journals/corr/abs-2103-14030,DBLP:journals/corr/abs-2010-04159,DBLP:conf/eccv/CarionMSUKZ20}, with no flexible or efficient way to jointly absorb both operations in a single atomic module which has great potential in working as a computational primitive for visual understanding.

In this work, we explicitly address these obstacles and develop a novel atomic operator, named \emph{X-volution}, integrating both convolution and self-attention.
First, we theoretically prove the feasibility of approximating the global self-attention by propagating contextual relationship from local region to non-local region.
According to this idea, we develop a novel approximate self-attention scheme with the complexity of $O(n)$, named \emph{Pixel Shift Self-Attention (PSSA)}.
In contrast to concurrent self-attention schemes~\cite{DBLP:conf/nips/VaswaniSPUJGKP17,DBLP:journals/corr/abs-2103-14030,DBLP:journals/corr/abs-2103-12731}, the proposed PSSA 
converts the self-attention into a convolutional operation on transformed features which are obtained by sequential element-wise shift and element-wise dot product. 
Second, based on this approximation scheme, we establish a multi-branch network module to integrate convolution and self-attention simultaneously.
The multi-branch topology makes the module not only possess well-conditioned optimization properties, but also acquire the ability to capture long-term relationships, thus showing stronger learning ability and better performance.
More importantly, 
our carefully designed multi-branch structure could be conditionally transformed into a single standard convolution operation via network structural re-parameterization~\cite{DBLP:journals/corr/abs-2101-03697}, rendering a pure convolution styled operator \emph{X-volution}, ready to be plugged into any modern networks as an atomic operation.
In fact, convolution and self-attention can be regarded as static convolution (\emph{i.e.}, content-independent kernel) and dynamic convolution (\emph{i.e.}, content-dependent kernel), respectively.



We experiment with the proposed X-volution in terms of both qualitative and quantitative evaluations on mainstream vision tasks.
Extensive results demonstrate that our X-volution operator achieves very competitive improvements (\emph{i.e.}, image classification on ImageNet~\cite{DBLP:journals/ijcv/RussakovskyDSKS15}: +1.2\% top1 accuracy, object detection on COCO~\cite{DBLP:conf/eccv/LinMBHPRDZ14}: +1.7 box AP, instance segmentation on COCO: +1.5 mask AP).

\section{Related Work}
Convolutional networks~\cite{DBLP:journals/corr/SimonyanZ14a,DBLP:conf/cvpr/HeZRS16,DBLP:conf/cvpr/SzegedyVISW16} have been proved to be powerful when facing computer vision problems such as object detection and image classification.
However, recent works~\cite{DBLP:conf/cvpr/0004GGH18} show that the effectiveness of the convolution operator is limited to its local receptive field and the self-attention mechanism with global information aggregation can achieve better performance.
Driven by this understanding, a lot of works~\cite{DBLP:conf/cvpr/0004GGH18,DBLP:conf/cvpr/HuSS18,DBLP:conf/cvpr/WangWZLZH20,DBLP:conf/eccv/WooPLK18} have introduced various complex mechanisms to enhance the representation ability of the network.
Recently, researchers realized that self-attention~\cite{DBLP:conf/nips/VaswaniSPUJGKP17} as a computational primitive can also handle computer vision problems with high performance, and a large number of transformers~\cite{DBLP:journals/corr/abs-2103-12731,DBLP:journals/corr/abs-2103-14030,DBLP:journals/corr/abs-2010-04159,DBLP:conf/eccv/CarionMSUKZ20,DBLP:journals/corr/abs-2101-11605,DBLP:journals/corr/abs-2103-15436,DBLP:journals/corr/abs-2103-16553,DBLP:journals/corr/abs-2012-09760} emerge.
Cordonnier et al.~\cite{DBLP:conf/iclr/CordonnierLJ20} found that in fact, multi-head self-attention 
is able to learn the characteristics of convolution, and the two can be converted to each other under certain conditions.
Instead of using convolution alone, Bello et al.~\cite{DBLP:conf/iccv/BelloZLVS19} combined convolution and self-attention through direct concatenation, and achieved a promising improvement.
This shows that combining the two operators is of great help in improving performance.
Wu et al.~\cite{DBLP:journals/corr/abs-2103-15808} confirmed the above point from another angle.
They introduced the convolution operation to vision transformers and further improved performance.



\section{Methodology}
We propose a novel atomic operator, named \emph{X-volution}, to integrate the fundamental convolution and self-attention operators into a single unified computing block, which is expected to gain very impressive performance improvements from both, \emph{i.e.,} local vs. non-local/linear vs. non-linear. In this section, we first revisit basic mathematical formulas of convolution and self-attention, and then we propose a simple approximation scheme of global self-attention, which is directly converted to a compatible pattern of convolution.
Finally, we describe that in the inference phase how to conditionally merge the branches of convolution and the proposed self-attention approximation into a \textbf{SINGLE} convolutional style atomic operator.

\begin{figure}[t]	
    \centering	
	\includegraphics[width=1\textwidth]{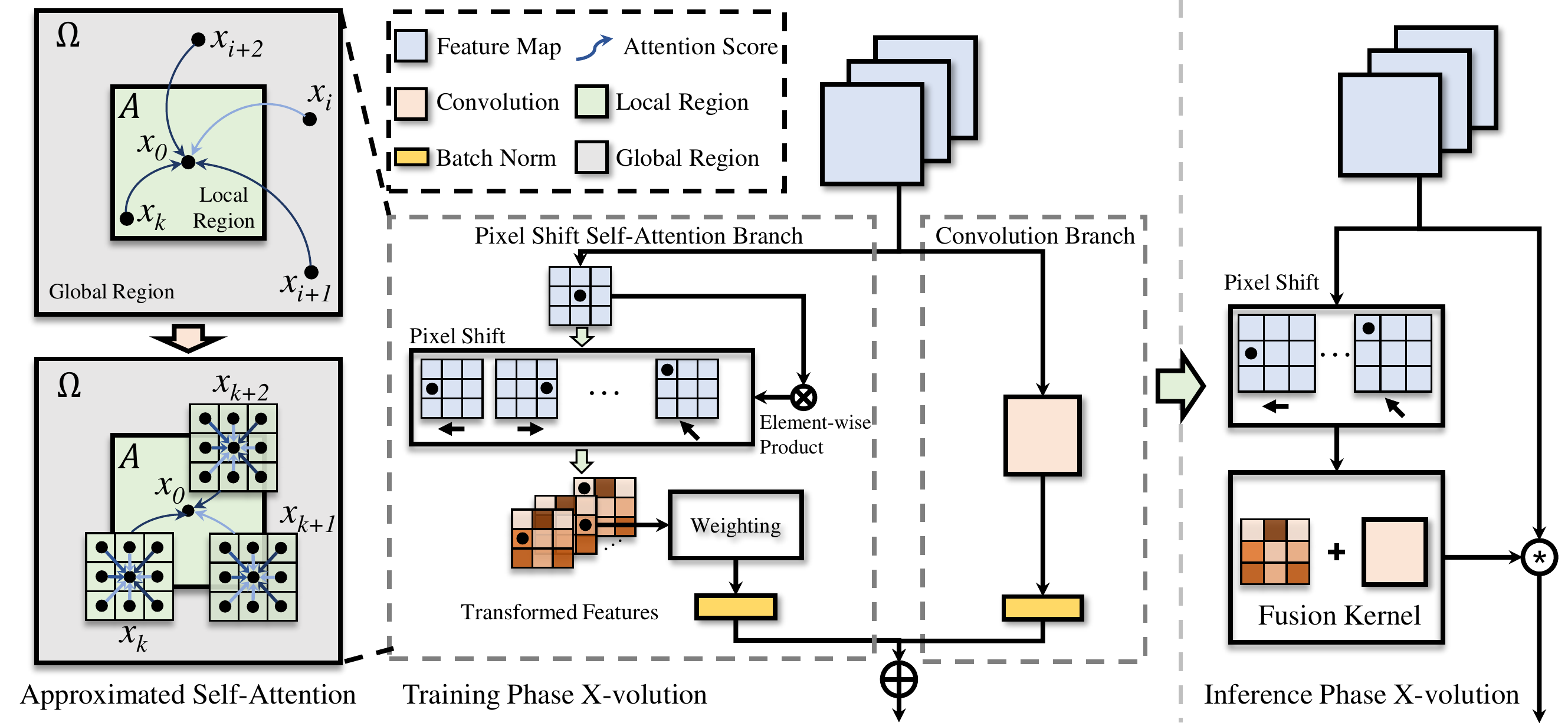}
	\small
	\caption{The detailed structure of our proposed X-volution operator. X-volution is a training-inference decouple topology~\cite{DBLP:journals/corr/abs-2101-03697}.
	Its training structure is shown in the middle with two main branches.
	Right branch is composed of cascaded convolution and BN, which can integrate convolution capabilities for X-volution.
	The left consists of a pixel shift self-attention, which provides approximated global self-attention features.
	Once trained, we conditionally re-parameterize it as an atomic operation as shown on the right.
	At inference phase, X-volution is actually a dynamic convolution operator, and its weight is composed of the attention map that needs to be dynamically calculated and the convolution weight that has been trained and solidified.
	}
    \label{fig:framework}
\end{figure}

\subsection{Convolution and Self-Attention Operator Revisit}

\subsubsection{The Convolution Module}
The convolutional operator is the de-factor computational primitive for building convolutional neural network (CNN), which estimates the output by linear weighting within a limited local region.
Given a feature tensor {\small$\mathbf{X}\in\mathbb{R}^{C_{i}\times H\times W}$}, {\small$C_{i}$} denotes the number of input channels, $H$ is the height and $W$ is the width.
The estimation result {\small$\mathbf{Y}\in\mathbb{R}^{C_{o}\times H\times W}$} of a convolutional operator is defined by following formulation:
\begin{equation}
\centering
\mathbf{Y}_{c_o,i,j} \overset{\underset{\mathrm{def}}{}}{=} \sum_{c_i=0}^{C_i}\sum_{(\delta_i,\delta_j)\in \mathbf{\Delta}_K} \mathcal{W}_{c_o,c_i,\delta_i+\lfloor K/2 \rfloor, \delta_j + \lfloor K/2 \rfloor} \mathbf{X}_{c_i,i+\delta_i,j+\delta_j} + \mathcal{B}_{c_o},
\label{eq:conv}
\end{equation}
where {\small$C_{o}$} denotes the number of output channels.
{\small$\mathcal{W}\in\mathbb{R}^{C_{o}\times C_{i}\times K\times K}$} refers to the convolution kernel, and {\small$\mathcal{W}_{c_o,c_i,\delta_i+\lfloor K/2 \rfloor, \delta_j + \lfloor K/2 \rfloor}$} refers to the kernel scalar value at the specific location. $K$ is the window size of convolution, {\small$\mathcal{B}\in \mathbb{R}^{C_{o}}$} denotes the bias vector, and {\small$\mathbf{\Delta}_K\in \mathbb{Z}^2$} represents the set of all possible offsets in the {\small$K \times K$} convolution window.
It can be seen from Eq.~\ref{eq:conv} that convolution is actually a first-order linear weighting operation.

\subsection{The Self-Attention Module}
The self-attention~\cite{DBLP:conf/nips/VaswaniSPUJGKP17} is a rising alternative computational primitive for vision tasks~\cite{DBLP:journals/corr/abs-2103-12731,DBLP:journals/corr/abs-2103-14030,DBLP:journals/corr/abs-2010-04159,DBLP:conf/eccv/CarionMSUKZ20}, whose core idea is to build long-term semantic interactions established by performing the intra-vector dot-product.
Unlike convolution, self-attention can not process image tensor directly, where input feature tensor is firstly reshaped into a vector {\small$\mathbf{X}\in \mathbb{R}^{C\times L}$}. $L$ represents the length of the vector and {\small$L = H\times W$}.
{\small$\mathcal{W}^Q$,
$\mathcal{W}^K$,
$\mathcal{W}^V$} respectively represent the Query, Key and Value embedding transformation,
and they are spatial-shared linear transformations.
The prediction of self-attention is defined as below:
\begin{equation}
\centering
\mathbf{Y} \overset{\underset{\mathrm{def}}{}}{=}
softmax\left(\left(\mathcal{W}^Q\mathbf{X}\right)^T\mathcal{W}^K\mathbf{X}\right)\mathcal{W}^V \mathbf{X}
= \Bar{\mathcal{W}}(\mathbf{X})\mathbf{X},
\label{eq:attention1}
\end{equation}
where {\small$\Bar{\mathcal{W}}(\mathbf{X})$} represents the final equivalent coefficient matrix of self-attention, which can be considered as a dynamic (element-wise content-dependent) and spatially varying convolutional filter.
Eq.~\ref{eq:attention1} demonstrates that self-attention is a high-order global operation.

\subsection{The Approximation of Global Self-Attention}
The global self-attention is the most primitive attention scheme, which possesses the advantage of excellent performance benefited from its global scope.
However, its complexity is unacceptable \emph{i.e.}, $O(n^2)$ ($n$ denotes total pixel number), which makes its application in visual tasks severely limited.
The key problem becomes whether we could infer a proper approximate scheme of {\small$\Bar{\mathcal{W}}(\mathbf{X})$} in Eq.~\ref{eq:attention1}, namely, whether we could find a compatible computing pattern of {\small$\Bar{\mathcal{W}}(\mathbf{X})$} in terms of off-the-shelf operators such as convolution, single element-wise product?
In this section, we show that after simple element-wise shift and dot-product, we could approximate the global self-attention operator in the form of convolution.
Given a position in the feature tensor $\mathbf{X}$, we denote its feature vector as $x_0$, and its attention logit $s_0$ can be written as following formulation:
\begin{equation}
\small
\centering
   s_0 = \sum_{x_t \in \Omega} \alpha_{t} \left \langle x_0,x_t\right\rangle
   = \underbrace{\sum_{x_j\in A} \alpha_j \left\langle x_0,x_j\right\rangle}_{Local\;Region} + \underbrace{\sum_{x_i\in (\Omega \setminus A)}\alpha_{i} \left\langle x_0,x_i\right\rangle}_{Non-local\;Region},
   \label{eq:attention_logit}
\end{equation}
where $\alpha_t = w^qw^kw^v x_t$, $\Omega$ denotes the global region, and $A$ denotes a local region centered on $x_0$.
We illustrate the local region and non-local region on the left side of Fig.~\ref{fig:framework}.
The gray box in the figure denotes the global region of input feature $\mathbf{X}$, and green box represents the local region centered on $x_0$.
In addition, non-local region is the region outside the local region.
Since images hold strong Markovian property~\cite{krahenbuhl2011efficient}, $x_0$ can be approximately linearly represented by pixels in its local region: {\small$x_0 \approx \sum_{x_k \in \mathring{A}} \beta_k x_k$},
where $\beta_k$ is the linear weight.
Substituting it into second item in Eq.~\ref{eq:attention_logit} can obtain following formulation:
\begin{equation}
\small
\sum_{x_i\in (\Omega\setminus A)}\alpha_{i} \left\langle x_0,x_i\right\rangle
\approx \sum_{x_i \in (\Omega\setminus A)} \alpha_i \left\langle \sum_{x_k\in \mathring{A}}\beta_k x_k,x_i\right\rangle = \sum_{x_i\in (\Omega\setminus A)} \sum_{x_k \in \mathring{A}}\alpha_i \beta_k \left\langle x_k,x_i\right\rangle.
\label{eq:second_item1}
\end{equation}
Without loss of generality, we can add terms in region $A$ whose coefficient is zero.
By design, non-local region is also within the receptive field of the boundary pixels of the local region.
Thus, we can transform Eq.~\ref{eq:second_item1} into below formulation:
\begin{equation}
\small
   \sum_{x_i\in (\Omega\setminus A)} \sum_{x_k \in \mathring{A}}\alpha_i \beta_k \left\langle x_k,x_i\right\rangle =  \sum_{x_i\in \Omega}\sum_{x_k \in \mathring{A}} \alpha_i \beta_k \left\langle x_k,x_i\right\rangle.
   \label{eq:111}
\end{equation}
According to the Markovian property of the image, we can assume that for $x_k\in A$, the interaction between $x_i$ (far away from $x_k$) and $x_k$ is weak.
Thus, the Eq.~\ref{eq:111} can be further simplified:
\begin{equation}
\small
 \sum_{x_i\in \Omega} \sum_{x_k \in \mathring{A}}\alpha_i \beta_k \left\langle x_k,x_i\right\rangle=
\sum_{x_k\in \mathring{A}}\sum_{x_i\in U(x_k)}\alpha_i \beta_k \left\langle x_k,x_i\right\rangle,
 \label{eq:xk}
\end{equation}
where $U(x_k)$ denote the local region of $x_k$.
Substituting Eq.~\ref{eq:xk} into second item in Eq.~\ref{eq:attention_logit}, we can rewrite it as following formulation:
\begin{equation}
\small
\begin{split}
\sum_{x_t \in \Omega} &\alpha_{t}  \left  \langle x_0,x_t\right\rangle 
\approx \sum_{x_i\in A} \alpha_i \cdot 1  \left\langle x_0,x_i\right\rangle + \sum_{x_k\in \mathring{A}}\sum_{x_i\in U(x_k)}\alpha_i \beta_k \left\langle x_k,x_i\right\rangle \\
&= \sum_{x_k\in A}\sum_{x_i\in U(x_k)} \alpha_i \beta_k \left\langle x_k,x_i\right\rangle
= \sum_{x_k\in A}  \beta_k \sum_{x_i\in U(x_k)} w^qw^kw^v x_i\left\langle x_k,x_i\right\rangle.
\label{eq:final_proof}
\end{split}
\end{equation}
Notice that, {\small$\left\langle x_k,x_i\right\rangle$} is the inner product between {\small$x_k$} and {\small$x_i$}, which  measures the similarity between {\small$x_k$} and {\small$x_i$}.
{\small$\sum_{x_i\in U(x_k)} \alpha_i \beta_k\left\langle x_k,x_i\right\rangle$} is the attention result of $x_k$ within its neighbourhood.
Therefore, the global attention logit at $x_0$ can be approximated by weighted summing the attention result of the pixels within its neighbourhood, as shown in Eq.~\ref{eq:final_proof}.
Following above understanding, we can design an approximate operator that estimates the global attention through point-by-point contextual relationship propagation.
Thus, we propose a global attention approximation scheme, \emph{Pixel Shift Self-Attention (PSSA)}, based on pixel shift and convolution to approximate the global attention.
Specifically, we first shift the feature map by $L$ pixels along given directions (\emph{i.e.}, left, right, up \emph{e.t.c.}), and then we employ element-wise product between the original features and the shifted features to obtain the transformed features.
In fact, the shift-product operation establishes the contextual relationship between the points in the neighborhood, and through hierarchical stacking, we can propagate the contextual relationship to the global region.
Finally, we perform weighted summing (can be implemented by convolution operator) between these transformed features to get an approximate self-attention map.
The complexity of shift, element-wise product and weighted summing are $O(n)$, so the proposed PSSA is an operator with $O(n)$ temporal complexity.
It is worth noting that PSSA actually converts self-attention into a standard convolution operation on transformed features.
Through hierarchical stacking, this structure can realize the estimation of global self-attention logit via contextual relationship propagation.


\subsection{The Unification of Convolution and Self-Attention: X-volution}
\subsubsection{Convolution and Self-Attention are Complementary}
The convolution employs the inductive bias of locality and isotropy endowing it the capability of translation equivariance~\cite{DBLP:conf/iccv/BelloZLVS19}.
However, the local inherent instinct makes convolution failed to establish the long-term relationship which is necessary to formulate a Turing complete atomic operator~\cite{DBLP:conf/iclr/CordonnierLJ20,DBLP:conf/iclr/PerezMB19}. 
In contrast to convolution, self-attention discards mentioned inductive bias, so-called \emph{low-bias}, and strives to discover natural patterns from a data-set without explicit model assumption.
The low-bias principle gives self-attention the freedom to explore complex relationships (\emph{e.g.}, long-term dependency, anisotropic semantics, strong local correlations in CNNs~\cite{DBLP:journals/corr/abs-2012-09841},  \emph{e.t.c.}), resulting in the scheme usually requires pre-training on extra oversize data-sets (\emph{e.g.}, JFT-300M, ImageNet21K).
In addition, self-attention is difficult to optimize, requiring a longer training cycle and complex tricks~\cite{DBLP:journals/corr/abs-2103-12731,DBLP:journals/corr/abs-2103-14030,DBLP:journals/corr/abs-2010-04159,DBLP:conf/eccv/CarionMSUKZ20}.
Witnessing this, several works~\cite{DBLP:conf/iccv/BelloZLVS19,DBLP:journals/corr/abs-2103-15808} propose that convolution should be introduced into the self-attention mechanism to improve its robustness and performance.
In short, different model assumptions are adopted to make convolution and self-attention complement each other in terms of optimization characteristics (\emph{i.e.}, well-condition/ill-condition), attention scope (\emph{i.e.}, local/long-term), and content dependence (content-dependent/independent) \emph{e.t.c.}.

\subsubsection{The Multi-Branch Topology for the Unification}

There are several works~\cite{DBLP:journals/corr/abs-2103-15808,DBLP:conf/iccv/BelloZLVS19} attempting to combine convolution and self-attention, whereas the coarse topological combination (\emph{e.g.}, hierarchical stacking, concatenate) prevents them from getting an single atomic operation (applying convolution and attention in the same module) and makes the structure irregular.
For instance, AANet~\cite{DBLP:conf/iccv/BelloZLVS19} directly concatenates the results processed by a convolution layer and a self-attention layer to obtain the combined results.
It shows that a single convolution or a single self-attention will cause performance degradation, and the performance will be significantly improved when they exist at the same time.

Although challenging due to heterogeneous computing pattern of convolution and self-attention, in this work, we study the mathematical formulation of convolution and self-attention operator, (\emph{i.e.}, Eq.~\ref{eq:conv} and Eq.~\ref{eq:attention1}), and find the approximate form in Eq.~\ref{eq:final_proof} could be equivalent to a spatial varying convolution on certain dot-product map, observing that global element-wise interaction (dot-product) could be approximated by the propagation of local element-wise interaction.
Thus, both operators could be treated in a unified computing pattern, \emph{i.e.}, convolution.
From the other point of view, the convolution operation can be regarded as the spatially invariant bias of the self-attention.
Perceiving this, we combine the operators into a \emph{multi-branch topology}, shown in the Fig.~\ref{fig:framework}, which could benefit from convolution and self-attention simultaneously.
The multi-branch module is composed of two main branches.
The branch on the left is composed of cascaded Shift Pixel Self-Attention and batch-normalization~\cite{DBLP:conf/icml/IoffeS15}, playing the role to approximate the global self-attention operation.
The right branch is designed as a convolutional branch composed of cascade convolution and batch-normalization.

\subsubsection{Conditionally Converting Multi-Branch Scheme to the Atomic X-volution}
The multi-branch module achieves the functional combination of convolution and self-attention.
However, it is only a coarse-grained combination of operators, which will make the network highly complex and irregular.
From the perspective of hardware implementation, the multi-branch structure requires more caches to serve the processing of multiple paths.
In contrast, a single atomic operation is more efficient and has lower memory overhead, which is hardware-friendly.
For brevity, we omit the formula of batch-normalization here.
Actually, batch-normalization can be regarded as a $1\times1$ group convolution (its group is equal to the number of channels), which can be merged into the convolution/self-attention layer.
In fact, we generally employ the PSSA by hierarchical stacking, and the weighted operation in the stacked structure can be omitted, as hierarchical stacking implies the operation of weighted neighbor pixels.
The training phase formulation of our proposed multi-branch module can be written as below:
\begin{equation}
y_0 = \underbrace{\sum_{x_i\in A}\alpha_i \left\langle x_0,x_i\right\rangle}_{PSSA\;Branch} + \underbrace{\sum_{x_i\in A} w^c x_i + b^{c}}_{Conv\;Branch}= \sum_{x_i\in A}w^q w^k w^v x_i\left\langle x_0,x_i\right\rangle + \sum_{x_i\in A} w^c x_i + b^{c},
\label{eq:attention}
\end{equation}
where $w^{c}$ represents the convolutional weight, and $b^{c}$ is its corresponding bias.
\begin{equation}
y_0 
=\sum_{x_i\in A}\left(w^q w^k w^v\left\langle x_0,x_i\right\rangle + w^c\right) x_i  + b^c = \sum_{x_i\in A}\left(w^{\mathcal{A}}(x_0,x_i)+w^c\right)x_i + b^c,
\label{eq:final_conv}
\end{equation}
where {\small{$w^{\mathcal{A}}(x_0,x_i)\overset{\underset{\mathrm{def}}{}}{=} w^q w^k w^v\left\langle x_0,x_i\right\rangle$}} represents the content-dependent/dynamic coefficients coming from pixel shift self-attention branch.
$w^{c}$ denotes the content-independent/static coefficients inherited from convolutional branch, and it will be fixed once the training is completed.
Observing Eq.~\ref{eq:final_conv}, we can find that after a simple transformation the multi-branch structure can be transformed into a convolution form.
It is worth pointing out that the process is widely used in CNN and is called structural re-parameterization~\cite{DBLP:journals/corr/abs-2101-03697}.
We here first extend it to the merging of convolution and self-attention.
According to Eq.~\ref{eq:final_conv}, we equivalently convert the multi-branch module composed of convolution and self-attention into a dynamic convolution operator named \emph{X-voultion}.
Note that, our proposed X-volution can be plugged in mainstream networks (\emph{e.g.}, ResNet) as an atomic operation.





\section{Experiments}
\subsection{Implementation Details}
\label{sec_4_1}
For image classification, we test the proposed X-volution on ImageNet-1k~\cite{DBLP:journals/ijcv/RussakovskyDSKS15} (MIT License) benchmark, which contains 1.28M training images and 50K validation images. Input images are cropped to 224 $\times$ 224 pixels with horizontal flipping. We use the SGD optimizer (initial learning rate: 0.1, momentum: 0.9, weight decay: $1 \times 10^{-4}$) with a total batch size of 256 to train the network for 100 epochs, and the learning rate is decreased by the factor of 10 every 30 epochs. 

For object detection and instance segmentation, We conduct experiments on COCO 2017~\cite{DBLP:conf/eccv/LinMBHPRDZ14} (Commons Attribution 4.0 License) data-set, which contains 118k training images and 5k validation images. 
We adopt Faster R-CNN~\cite{DBLP:conf/nips/RenHGS15} for detection and Mask R-CNN~\cite{DBLP:conf/iccv/HeGDG17} for segmentation, both equipped with the FPN~\cite{DBLP:conf/cvpr/LinDGHHB17} neck. 
We employ the implementation of detectors from Detectron2~\cite{wu2019detectron2} with its default settings.
Concretely, a total batch size of 16 (2 images per GPU) is adopted to train the network with \textbf{1x} (12 epochs) schedule, and the default multi-scale training strategy is utilized. 

For all experiments, we choose ResNet~\cite{DBLP:conf/cvpr/HeZRS16} as the backbone. Following BoTNet~\cite{DBLP:journals/corr/abs-2101-11605}, we replace the final three $3 \times 3$ convolutions with the proposed X-volution to verify its effectiveness, and details of operator replacement are shown in the Fig.~\ref{fig:arch}. The detailed implementation of PSSA can be found in the supplementary material. Unless otherwise specified, the replacement only occurs in the last stage of ResNet, and other operators are evaluated with the same replacement for fair comparison. 
In specific, we re-evaluate the performance of AA-Convolution (presented by AA-Net~\cite{DBLP:conf/iccv/BelloZLVS19}), Self-Attention (SA)~\cite{DBLP:conf/nips/VaswaniSPUJGKP17}, and Involution~\cite{DBLP:journals/corr/abs-2103-06255} operators using the implementation from their source code under the same setting, and the results of our X-volution combined with other self-attention operators (X-volution(SA), X-volution(Inv)) are also provided. The reported results are averaged over five independent runs.
We use 8 NVIDIA Tesla V100 GPUs for training. 
All the operators are implemented with the PyTorch~\cite{DBLP:conf/nips/PaszkeGMLBCKLGA19} deep learning framework, and the source code will be released for reproducibility. 
Note that, no other data augmentation or training skills are adopted.

\vspace{-1mm}
\subsection{Main Results}

\subsubsection{Image Classification}

\begin{figure}[t]
    \centering
    \small
    \begin{minipage}[t!]{0.47\textwidth}
        \centering
        \makeatletter\def\@captype{table}\makeatother
        \caption{Comparison of different operators on ImageNet-1K~\cite{DBLP:conf/cvpr/DengDSLL009} using ResNet-34 and ResNet-50.
        }
        \begin{spacing}{1.05}
         \small
            \setlength\tabcolsep{4.5pt}
                \begin{tabular}{c| l| c| c}
                \Xhline{1.0pt}
        		Backbone &  Operator & top-1 & top-5
        		\\ \hline
        		
        		\multirow{6}{*}{ResNet-34} 
        		& Convolution & 73.8 & 91.6\\ 
                & Self-Attention & 73.2 & 91.4 \\
                & PSSA & 73.6 & 91.5 \\
                \cline{2-4}
        		& X-volution, \textit{stage3} & \textbf{75.0} & \textbf{92.4}  \\
        		& X-volution, \textit{stage4} & 74.3 & 91.9 \\
        		& X-volution, \textit{stage5} & 74.2 & 91.7 \\
        		\hline
        		
        		\multirow{6}{*}{ResNet-50} 
        		& Convolution & 75.7 & 92.5 \\ 
                & Self-Attention & 75.3 & 92.2\\
                & PSSA & 75.5 & 92.5 \\
                \cline{2-4}
        		& X-volution, \textit{stage3} & \textbf{76.6} & \textbf{93.3}\\
        		& X-volution, \textit{stage4} & 75.1 & 92.4\\
        		& X-volution, \textit{stage5} & 75.9 & 92.8\\
                \Xhline{1.0pt}
        \end{tabular}
        
        \label{tab_cls}
        \vspace{-2mm}
        \end{spacing}
    \end{minipage}
    \hspace{5.0mm}
    \begin{minipage}[t!]{0.47\textwidth}
        \small
        \includegraphics[width=1.0\textwidth]{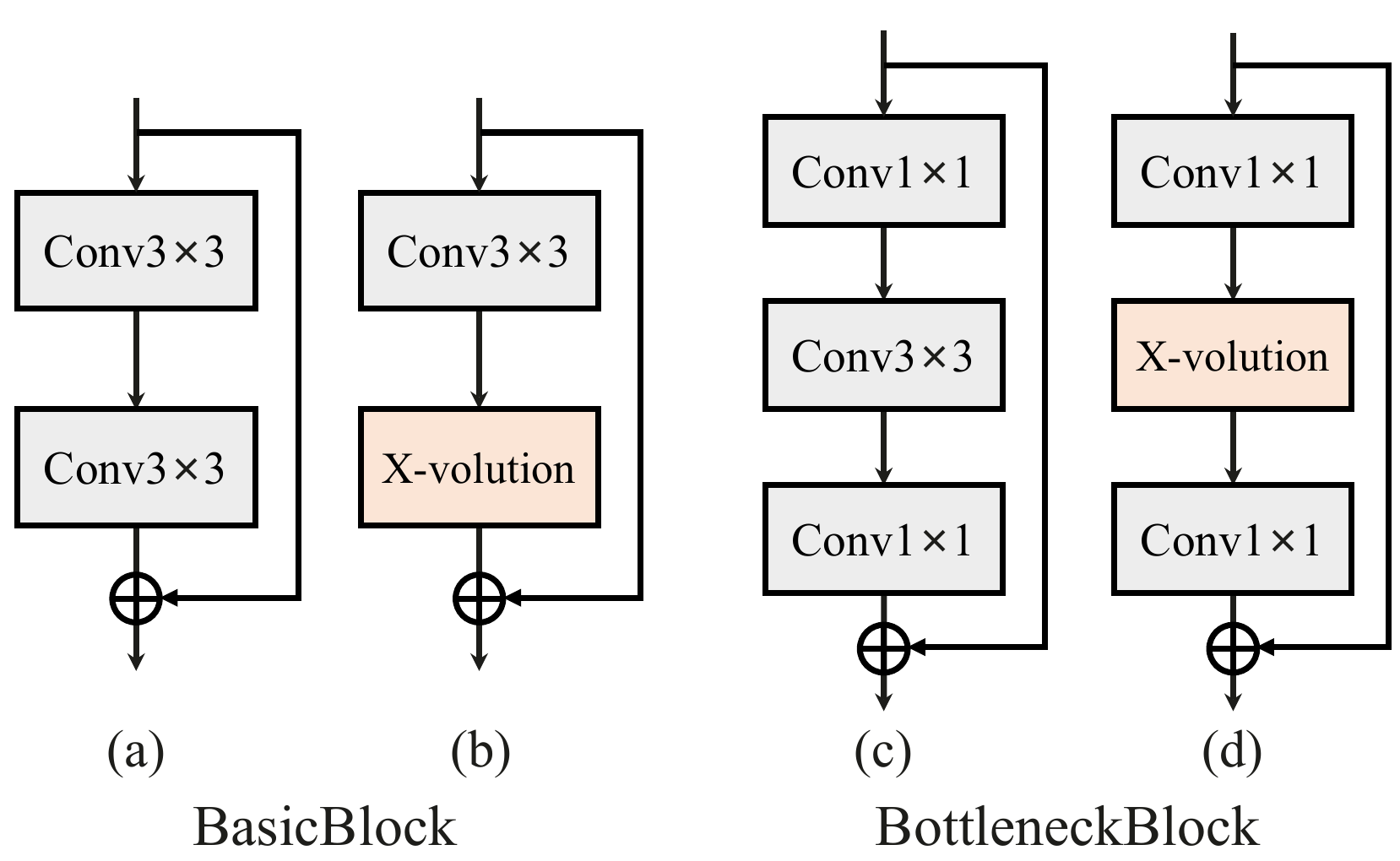}
        \caption{
        Replacement position for our X-volution operator. In ResNet-34, we replace the last \texttt{Conv3$\times$3} in the BasicBlock, and for ResNet-50, we replace the middle \texttt{Conv3$\times$3} in the BottleneckBlock.
        }
        \label{fig:arch}
    \end{minipage}
    \vspace{-2mm}
\end{figure}

\begin{table*}[t]\centering
 \caption{Object detection results of different operators on COCO val 2017.}
 \label{tab_det}
 \begin{spacing}{1.2}
 \small
 \setlength\tabcolsep{6.4pt}
 \begin{threeparttable}
 \begin{tabular}{l|c|c|c|c|c|c}
        \Xhline{1.1pt}
        Operator
		& \fontsize{8pt}{1em}\selectfont \quad\apbbox{~}
		& \fontsize{8pt}{1em}\selectfont \quad\apbbox{50}
		& \fontsize{8pt}{1em}\selectfont \quad\apbbox{75}
		& \fontsize{8pt}{1em}\selectfont \quad\apbbox{S}
		& \fontsize{8pt}{1em}\selectfont \quad\apbbox{M}
		& \fontsize{8pt}{1em}\selectfont \quad\apbbox{L} \\ \hline

		Convolution & 38.6 & 59.7 & 42.0 & 23.6 & 42.3 & 50.0 \\ 
		Multi-Convolution & 38.4\cgap{-}{0.2}  & 59.0\cgap{-}{0.7}  & 41.7\cgap{-}{0.3}  & 22.2\cgap{-}{1.4} & 41.6\cgap{-}{0.7} &  49.9\cgap{-}{0.1} \\
		AA-Convolution~\cite{DBLP:conf/iccv/BelloZLVS19}
		& 39.4\cgap{+}{0.8}  & 60.6\cgap{+}{0.9}  & 42.7\cgap{+}{0.7}  & 24.4\cgap{+}{0.8} &  42.9\cgap{+}{0.6} &  50.9\cgap{+}{0.9} \\ 
	
		\cline{1-7}
		
		Involution~\cite{DBLP:journals/corr/abs-2103-06255}
		& 38.8\cgap{+}{0.2} & 59.9\cgap{+}{0.2} & 42.1\cgap{+}{0.1} & 23.6\cgap{+}{0.0} & 41.9\cgap{-}{0.4} & 50.2\cgap{+}{0.2} \\ 
		X-volution(Inv) & 39.2\cgap{+}{0.6} & 60.2\cgap{+}{0.5} & 42.7\cgap{+}{0.7} & 23.6\cgap{+}{0.0} & 42.0\cgap{-}{0.3} & 
		50.9\cgap{+}{0.9}\\ \cline{1-7}
		
		Self-Attention~\cite{DBLP:conf/nips/VaswaniSPUJGKP17}
		&  38.5\cgap{-}{0.1} & 60.5\cgap{+}{0.8}  & 41.7\cgap{-}{0.3}  & 24.2\cgap{+}{0.6}  &  42.2\cgap{-}{0.1} & 48.6\cgap{-}{1.4}  \\ 
		X-volution(SA)& \textbf{40.3}\cgap{+}{1.7} & \textbf{61.8}\cgap{+}{2.1} & \textbf{43.7}\cgap{+}{1.7} & \textbf{24.6}\cgap{+}{1.0} & \textbf{43.9}\cgap{+}{1.6} & \textbf{52.0}\cgap{+}{2.0} \\
		\cline{1-7}
		
		PSSA & 38.7\cgap{+}{0.1} & 60.4\cgap{+}{0.7} & 42.2\cgap{+}{0.2} & 23.8\cgap{+}{0.2} & 42.3\cgap{+}{0.0} & 49.1\cgap{-}{0.9} \\ 
		X-volution & 40.1\cgap{+}{1.5} & 61.3\cgap{+}{1.6} & 43.6\cgap{+}{1.6} & 24.0\cgap{+}{0.4} & 43.6\cgap{+}{1.3} & 51.4\cgap{+}{1.4} \\
		\Xhline{1.0pt}
		
\end{tabular}
	\begin{tablenotes}
    \footnotesize
    \item[*] $\pm(x.x)$ here denotes the performance gain over the convolution baseline.
	\end{tablenotes}
\vspace{-3mm}
\end{threeparttable}
\end{spacing}
\end{table*}

In Tab.~\ref{tab_cls}, we compare our proposed X-volution operator with the fundamental convolution and self-attention operator using ResNet-34 and ResNet-50 as the backbone on ImageNet-1k. 
We replace partial convolution operators with our proposed X-volution in the network, and the detailed replacement position is shown in Fig.~\ref{fig:arch}.
We can observe that our proposed X-volution leads to stable improvement for both ResNet structures, which illustrates the effectiveness of our multi-branch design. The self-attention branch provides our X-volution with non-local and anisotropic processing capabilities, which are very important for dealing with complex context relationships.
Note that, the performance of stand-alone self-attention operator inserted into ResNet is worse than convolution, indicating that the naive introduction of self-attention operator has little effect on classification task.

We also investigate the effect of different replacement locations. 
Different stages have different numbers of building blocks, \emph{e.g.}, 3, 4, 6, 3 for ResNet-34 and ResNet-50, and also correspond to different feature resolution. 
Limited by computational resources, 
we only test our X-volution in the last three stages, replacing all residual blocks in this stage. 
Results show that the replacement in \textit{stage3} brings greatest improvement, \emph{i.e.}, +1.2\% top-1 accuracy for ResNet-34 and +0.9\% top-1 accuracy for ResNet-50. 
We suspect the inferior performance of the replacement in \textit{stage4} for ResNet-50 can be ascribed to the increased learnable parameters, which slow down the convergence of the network.


\subsubsection{Object Detection}
Besides image classification, we also evaluate our proposed X-volution operator on object detection to verify its generalization ability. Tab.~\ref{tab_det} reports the results of different operators on COCO 2017. 
Convolution (first row) denotes the original ResNet-50 baseline trained with ImageNet pre-trained weights. 
The performance of all operators is obtained by only replacing the final three 3$\times$3 convolutions of the last stage in ResNet-50 architecture, as shown in Fig.~\ref{fig:arch} (d). 

Based on the Faster R-CNN framework, three versions of X-volution all surpass their self-attention counterparts with considerable performance gains.
In particular, our X-volution(SA) achieves the best performance with a significant gain of +1.7 box AP over ResNet-50.
By combining low-order local features and high-order long-range dependencies, the proposed X-volution operator achieves higher accuracy than convolution or self-attention alone.
The results show that, a Turing complete atomic operator is helpful to visual understanding, and such property is neglected by existing computational operators.
Moreover, the X-volution based on PSSA achieves comparable performance with the X-volution(SA), indicating that the approximation works well in our X-volution module, which is more friendly to hardware implementation and computation.
We also verified the effect of multi-branch convolution. Under the same training settings, the design of multi-branch convolution (second row in Tab.~\ref{tab_det}) leads to degraded performance, showing that the increase of network learnable parameters does not always lead to performance improvement.




\begin{table*}[t]\centering
 \caption{Instance segmentation results of different operators on COCO val 2017.}
 \begin{spacing}{1.3}
 \small
 \setlength\tabcolsep{0.9pt}
 \begin{tabular}{l|c|c|c|c|c|c|c|c}
        \Xhline{1.1pt}
		Operator
		& \fontsize{7pt}{1em}\selectfont \quad\apbbox{~}
		& \fontsize{7pt}{1em}\selectfont \quad\apbbox{S}
		& \fontsize{7pt}{1em}\selectfont \quad\apbbox{M}
		& \fontsize{7pt}{1em}\selectfont \quad\apbbox{L}
		& \fontsize{7pt}{1em}\selectfont \quad\apmask{~}
		& \fontsize{7pt}{1em}\selectfont \quad\apmask{S}
		& \fontsize{7pt}{1em}\selectfont \quad\apmask{M}
		& \fontsize{7pt}{1em}\selectfont \quad\apmask{L}\\ \hline
        
        Convolution &39.1 & 23.2 & 42.4 & 51.2 & 35.7 & 17.4 & 38.1 & 51.4 \\
		AA-Convolution~\cite{DBLP:conf/iccv/BelloZLVS19}
		& 40.2\cgap{+}{1.1} & 25.0\cgap{+}{1.8} & 43.6\cgap{+}{1.2} & 52.0\cgap{+}{0.8} & 36.4\cgap{+}{0.7} & 18.8\cgap{+}{1.4} & 39.2\cgap{+}{1.1} & 52.5\cgap{+}{1.1} \\
		\cline{1-9}

		Involution~\cite{DBLP:journals/corr/abs-2103-06255}
		& 39.6\cgap{+}{0.5} & 23.7\cgap{+}{0.5} & 42.9\cgap{+}{0.5} & 51.4\cgap{+}{0.2} & 35.8\cgap{+}{0.1} & 18.0\cgap{+}{0.6} & 38.4\cgap{+}{0.3} & 51.3\cgap{-}{0.1} \\
		X-volution(Inv) &  39.9\cgap{+}{0.8} & 24.1\cgap{+}{0.9} & 42.9\cgap{+}{0.5} & 52.0\cgap{+}{0.8} & 35.9\cgap{+}{0.2} & 18.2\cgap{+}{0.8} & 38.5\cgap{+}{0.4} & 51.8\cgap{+}{0.4}\\
		\cline{1-9}
		   
		Self-Attention~\cite{DBLP:conf/nips/VaswaniSPUJGKP17}
		& 39.4\cgap{+}{0.3} & 24.1\cgap{+}{0.9} & 43.3\cgap{+}{0.9} & 49.8\cgap{-}{1.4} & 36.1\cgap{+}{0.4} & 18.1\cgap{+}{0.7} & 39.0\cgap{+}{0.9} & 51.0\cgap{-}{0.4} \\
		X-volution(SA) & \textbf{41.1}\cgap{+}{2.0} & \textbf{25.4}\cgap{+}{2.2} & \textbf{44.3}\cgap{+}{1.9} & 53.0\cgap{+}{1.8} & \textbf{37.2}\cgap{+}{1.5} & \textbf{19.2}\cgap{+}{1.8} & \textbf{40.0}\cgap{+}{1.9} & \textbf{53.1}\cgap{+}{1.7}\\ 
		\cline{1-9}
	
		PSSA & 39.5\cgap{+}{0.4} & 23.4\cgap{+}{0.2} & 43.1\cgap{+}{0.7} & 50.7\cgap{-}{0.5} & 36.0\cgap{+}{0.3} & 17.1\cgap{-}{0.3} & 38.6\cgap{+}{0.5} & 51.3\cgap{-}{0.1}\\
		X-volution &  40.9\cgap{+}{1.8} & 24.8\cgap{+}{1.6} & 43.8\cgap{+}{1.4} & \textbf{53.2}\cgap{+}{2.0} & 36.8\cgap{+}{1.1} & 18.5\cgap{+}{1.1} & 39.2\cgap{+}{1.1} & 52.7\cgap{+}{1.3} \\

		\Xhline{1.0pt}

\end{tabular}
\label{tab_seg}
\end{spacing}
\end{table*}

\subsubsection{Instance Segmentation}
For instance segmentation, we adopt Mask R-CNN framework with FPN neck and ResNet-50 backbone. Tab.~\ref{tab_seg} compares our X-volution operator against other atomic operators. We can observe that our proposed X-volution outperforms other operators by a large margin. Specifically, X-volution(SA) achieves 41.1 box AP and 37.2 mask AP, which brings 2.0 box AP and 1.5 mask AP gains over ResNet-50, and also improves Self-Attention by 1.7 box AP and 1.1 mask AP. The combination of the Involution~\cite{DBLP:journals/corr/abs-2103-06255} performs marginally better than the Involution baseline (+0.3 box AP and +0.1 mask AP), and the X-volution composed of PSSA contributes to comparable performance with the global version X-volution.
These results illustrate the superiority of our X-volution operator, in which the integration of local and global contextual information enables more precise instance segmentation prediction. 
Compared to AA-Convolution~\cite{DBLP:conf/iccv/BelloZLVS19}, which proposes a split attention mechanism, our multi-branch design of X-volution obtains a better performance, showing the complementarity and necessity of convolution and self-attention.




\subsection{Analysis Study}

\subsubsection{Effect of Different Kernel Size}
\begin{figure}[t]
    \centering
    \small
    \begin{minipage}[t!]{0.56\textwidth}
        \centering	
	\includegraphics[width=1.0\textwidth]{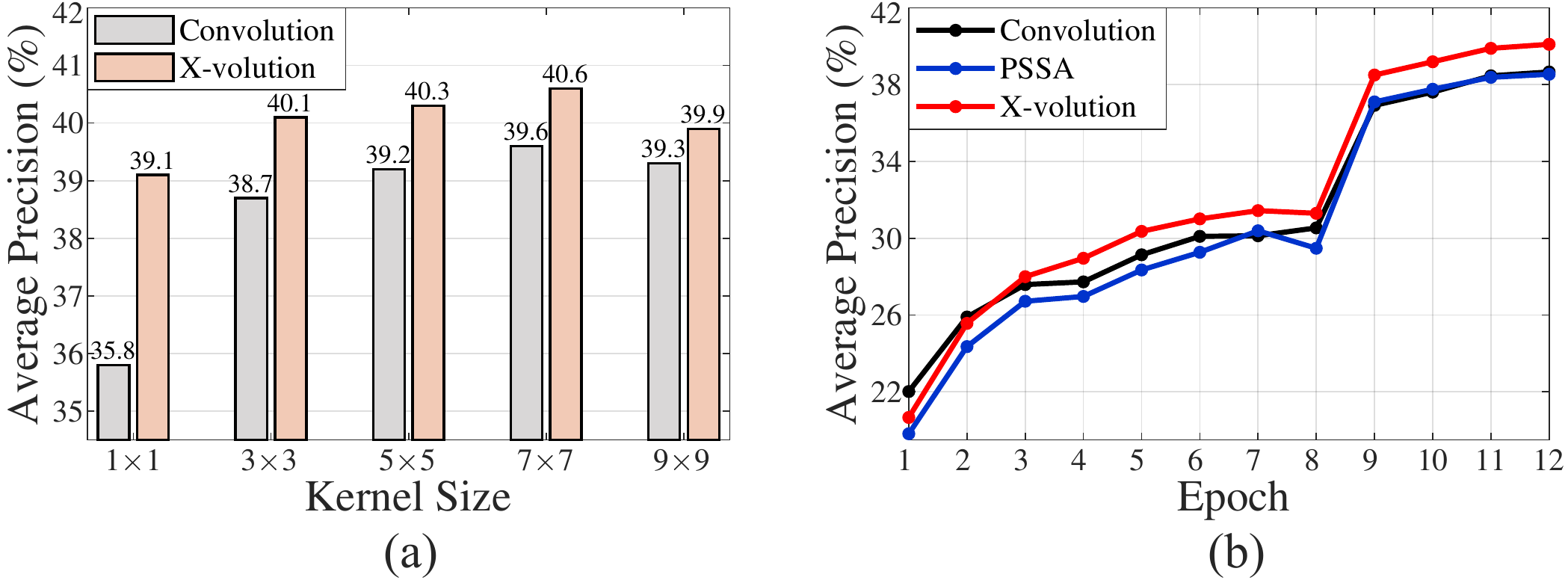}
	\vspace{-5mm}
	\caption{(a) Ablation study for different convolution kernel sizes.(b) Convergence speed of different kinds of operators on COCO (learning rate decays at $8$-th and $11$-st epoch).}
    \label{fig:kernel}
    \end{minipage}
    \hspace{4.0mm}
    \begin{minipage}[t!]{0.40\textwidth}
        \centering
        \makeatletter\def\@captype{table}\makeatother
        \caption{Effects of relative position encoding on COCO detection. Following~\cite{DBLP:journals/corr/abs-2103-14030}, we add the position embedding as the bias to the product of Query and Key.}
        \vspace{0.5mm}
        \label{tab:position encoding}
        \small
        \begin{spacing}{1.25}
        \setlength{\tabcolsep}{0.4pt}
        \begin{tabular}{c| c| c c c c}
        \Xhline{1.0pt}

        Operator & Pos. 
        & \fontsize{6.5pt}{1em}\selectfont \quad\apbbox{~}
        & \fontsize{6.5pt}{1em}\selectfont \quad\apbbox{S}
		& \fontsize{6.5pt}{1em}\selectfont \quad\apbbox{M}
		& \fontsize{6.5pt}{1em}\selectfont \quad\apbbox{L} \\
        \hline
        
        PSSA &  & 38.7 & 23.8 & 42.3 & 49.1\\
        PSSA & \checkmark & 39.1 & 24.1 & 42.7 & 49.5 \\
        \hline
        X-volution &  & 40.1 & 24.0 & 43.6 & 51.4 \\
        X-volution & \checkmark & 40.2 & 24.0 & 43.8 & 51.7 \\
       \Xhline{1.0pt}
        \end{tabular}
        \end{spacing}
    \end{minipage}
\end{figure}

To study the influence of convolution branch in our X-volution, we conduct experiments of different convolutional kernel size (\emph{i.e.}, ranges from $1\times 1$ to $9 \times 9$).
We show the results in Fig.~\ref{fig:kernel} (a).
When reducing the kernel size to $1\times 1$, obvious performance decay occurs, which is possibly related to the feature resolution in the network. 
On the contrary, the performance of X-volution under this configuration is still acceptable, as self-attention branch provides the ability to build long-term interactions.
We continue to increase the size of the convolution kernel (\emph{i.e.}, $5\times 5$, $7\times 7$), and steady improvements are observed on the corresponding network. 
When the size of the convolution kernel is set to $9\times 9$, the performance begins to decrease.
It is worth noting that the improvement of X-volution over stand-alone convolution becomes smaller.
This phenomenon is caused by two factors.
First, the large kernel size increases the number of learnable parameters, which inevitably improves the capability of the network and thus brings the improvement.
Second, increasing the size of the convolution kernel will enlarge the receptive field and gain the ability to establish relationship in a larger region.
It can be concluded from this experiment that the convolution branch plays a critical role in our X-volution and the existence of the self-attention branch allows X-volution to achieve an excellent competitive performance.

\subsubsection{The Optimization Property of X-volution}\label{sec_optimization}
Convolution obtains well-conditioned optimization characteristics due to its local and isotropic inductive bias, but self-attention is a low-bias operator, which makes its optimization difficult and requires more training epochs.
We show the AP convergence curve of different operators on COCO object detection in Fig.~\ref{fig:kernel} (b). In the initial stage (\emph{i.e.}, the $1$-st epoch), the performance of convolution is obviously higher than that of X-volution, and the convergence speed of our X-volution is slightly faster than the single PSSA branch thanks to the addition of convolution.
As training continues, our X-volution surpasses convolution and PSSA in the $3$-th epoch, and maintains the lead in performance.
Our designed PSSA lags behind convolution until the $9$-th epoch, at which time the learning rate is decayed, and then these two operators maintain comparable performance.
From the trend of the curve in Fig.~\ref{fig:kernel}, it can be concluded that combining the two operators can overcome the shortcoming of slow convergence of self-attention and obtain better optimization properties.

\vspace{2mm}
\subsubsection{Relative Position Encoding}
Positional encoding\cite{DBLP:conf/nips/VaswaniSPUJGKP17, DBLP:conf/naacl/ShawUV18} is an vital component in self-attention.
When processing, self-attention needs to transform the data into a vector, which destroys the position information in the data.
This is particularly prominent in visual tasks.
Contemporary transformers\cite{DBLP:journals/corr/abs-2103-14030, DBLP:journals/corr/abs-2101-11605, DBLP:conf/eccv/CarionMSUKZ20} employ various position embedding methods (\emph{e.g.}, absolute position encoding, relative position encoding, \emph{e.t.c.}) to recover the perception of positional information.
In Tab.~\ref{tab:position encoding}, we report the results of ablation study for relative position encoding.
It can be observed that, for our X-volution, there is only a marginal improvement after adding the position encoding, compared to the approximate self-attention.
We judge that the local processing properties of convolution can alleviate the problem of missing position information caused by self-attention to a certain extent.
Moreover, relative position encoding possesses similar characteristics with local processing, which is directly introduced by the convolution branch in our X-volution.
Based on this understanding, the standard X-volution is not equipped with position encoding, which also slightly reduces additional computational burden.

\vspace{2mm}
\section{Conclusion and Future Work}
\label{sec_5}
In this paper we study and analyze the fundamental principle of convolution and self-attention, which show complementary characteristics.
We first theoretically derive a global self-attention approximation scheme PSSA. 
Then we propose a multi-branch topology to integrate the two operators in a coarse-grained manner, absorbing the advantages from both.
Furthermore, we leverage the structural re-parameterization to perform fine-grained merging of the constructed multi-branch after training, and finally obtain a single convolutional style atomic operator X-volution which simplifies the topology.
Extensive experiments on image classification, object detection and instance segmentation demonstrate the effectiveness and feasibility of our proposed operator.

Several opening problems for this work still remain.
In future work, we will focus on how to further reduce the complexity of operator merging.
Besides, the experiments employing the X-volution in various backbone should be conducted, which could further demonstrate the feasibly and efficiency of our operator.
In addition, we will study and design a framework entirely composed of X-volution.


\clearpage
\section{Appendix}
\subsection{More Implementation Details} 
\label{supp_sec1}
\subsubsection{Network Architecture}
In Tab.~\ref{tab_arch}, we show the detailed architecture we used for our X-voluiton model (X-volution, \textit{stage5}) on ImageNet-1k classification. The only difference from ResNet-50 is replacement of 3$\times$3 convolutions with the proposed X-volution in \texttt{conv5\_x}. Other operators are evaluated with the same replacement.

\newcommand{\blockb}[3]{\multirow{3}{*}{
\(\left[
\begin{array}{l}
\text{1$\times$1, #2}\\
[-.2em] \text{3$\times$3, #2}\\
[-.2em] \text{1$\times$1, #1}
\end{array}\right]\)$\times$#3}
}

\newcommand{\blockBoT}[3]{\multirow{3}{*}{
\(\left[
\begin{array}{c}
\text{1$\times$1, #2}\\
[-.2em] \text{{\textcolor{myred}{\emph{X-volution}}}, #2}\\
[-.2em] \text{1$\times$1, #1}
\end{array}\right]\)$\times$#3}
}

\newcolumntype{x}[1]{>\centering p{#1pt}}
\newcommand{\ft}[1]{\fontsize{#1pt}{1em}\selectfont}
\renewcommand\arraystretch{1.20}
\setlength{\tabcolsep}{4.0pt}
\begin{table}[!htb]
\caption{Architectures for ImageNet using ResNet-50.}
\begin{center}
\small
\begin{tabular}{c|c|c|c}
\Xhline{1.0pt}
 stage & output & ResNet-50 & \textbf{Ours} \\
\hline
\texttt{conv1} & 112$\times$112 & \multicolumn{2}{c}{7$\times$7, 64, stride 2}\\
\hline
\multirow{4}{*}{\texttt{conv2\_x}} & \multirow{4}{*}{$56\times56$} &\multicolumn{2}{c}{3$\times$3 max pool, stride 2}
\\\cline{3-4}
  &  &  \blockb{256}{64}{3} & \blockb{256}{64}{3}\\
  &  &  & \\
  &  &  & \\
\hline

\multirow{3}{*}{\texttt{conv3\_x}} &  \multirow{3}{*}{$28\times28$} 
  & \blockb{512}{128}{4} &  \blockb{512}{128}{4}\\
  &  &  & \\
  &  &  & \\
\hline
\multirow{3}{*}{\texttt{conv4\_x}} & \multirow{3}{*}{$14\times14$} 
  & \blockb{1024}{256}{6} & \blockb{1024}{256}{6}\\
  &  &  & \\
  &  &  & \\
\hline
\multirow{3}{*}{\texttt{conv5\_x}} & \multirow{3}{*}{$7\times7$} 
& \blockb{2048}{512}{3} & \blockBoT{2048}{512}{3}\\
  &  &  & \\
  &  &  & \\
\hline
& {$1\times1$} & \multicolumn{2}{c}{average pool, 1000-d fc, softmax} \\
\Xhline{1.0pt}

\end{tabular}
\end{center}
\label{tab_arch}
\vspace{-.5em}
\end{table}

\subsubsection{Implementation of X-volution}

\begin{figure}[t]	
    \centering	
	\includegraphics[width=0.95\textwidth]{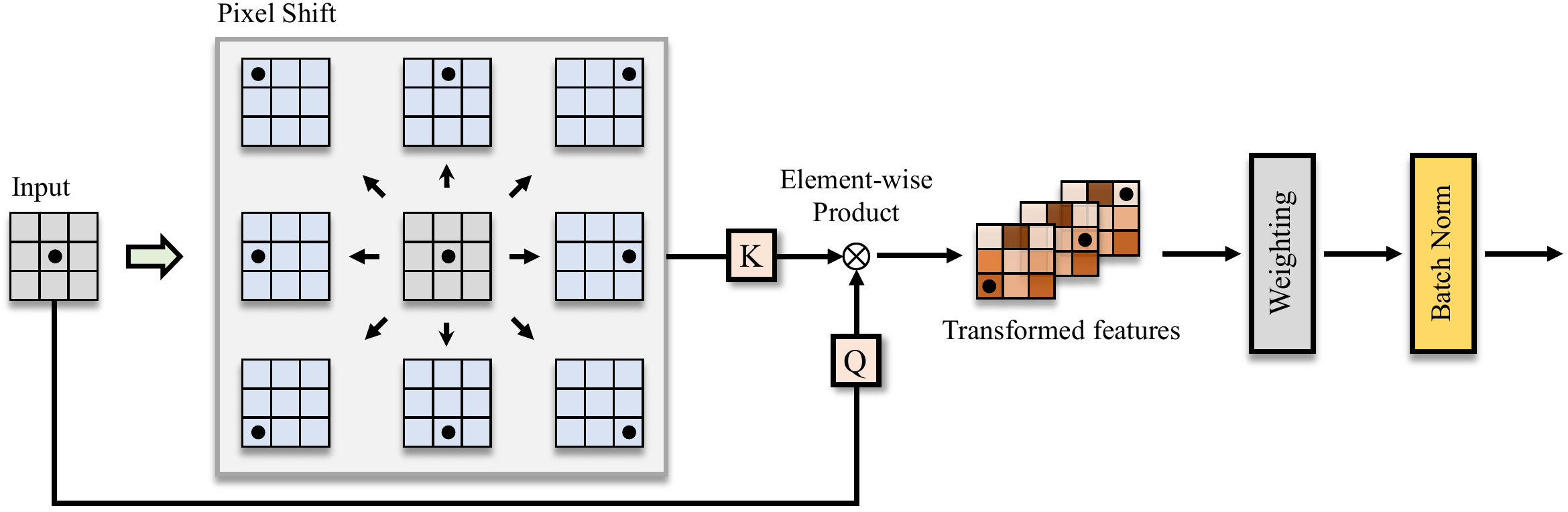}
	\small
	\caption{The detailed structure of our proposed PSSA operator.}
    \label{fig:pssa}
\end{figure}

Our proposed X-volution is composed of two branches, one of which is convolution and the other branch is composed of self-attention. For the convolution branch, we use a 3 $\times$ 3 convolution paralleled with an additional 5 $\times$ 5 dilated convolution (stride=1, dilation=2). For the self-attention branch, we use our proposed PSSA, which is an approximation of global self-attention, shown in Fig.\ref{fig:pssa}. In PSSA, the input feature map is firstly shifted toward eight directions by L pixels, where L is set as [1,3,5] in the implementation. After the Query and Key transformation, we calculate the element-wise products between the original feature map and the shifted feature map to get the transformed features. Then we perform weighted addition on the transformed features to get an approximated self-attention map. A batch normalization layer is stacked before the final output.

\subsection{More Detection Results on COCO}
\label{supp_sec2}

We also evaluate our proposed PSSA and X-volution with convolution baseline under multiple training schedules (\emph{i.e.}, \textbf{1x}: 12 epochs, \textbf{2x}: 24 epochs, \textbf{3x}: 36 epochs, \textbf{6x}: 72 epochs)\footnote{1x, 2x, 3x and 6x setting is adopted from Detectron2. \vspace{2mm}}.
For all experiments, we train with a total batch size of 16 on 8 NVIDIA Tesla V100 GPUs. The learning rate is initially set as 0.2, and is divided by 10 with different learning schedules. The other experimental settings remain the same. Tab.~\ref{tab:lr schedule} presents the detailed hyperparameters for different schedules.

\begin{table*}[!htb]\centering
 \caption{More detection results about different operators on COCO 2017 val under the multiple settings. 
}
 \label{tab:more det results}
 \begin{spacing}{1.05}
 \small
 \setlength\tabcolsep{5pt}
 \begin{threeparttable}
 \begin{tabular}{l|c|c|c|c|c|c|c}
        \Xhline{1.1pt}
        Operator
        & Epoch
		& \fontsize{8pt}{1em}\selectfont \quad\apbbox{~}
		& \fontsize{8pt}{1em}\selectfont \quad\apbbox{50}
		& \fontsize{8pt}{1em}\selectfont \quad\apbbox{75}
		& \fontsize{8pt}{1em}\selectfont \quad\apbbox{S}
		& \fontsize{8pt}{1em}\selectfont \quad\apbbox{M}
		& \fontsize{8pt}{1em}\selectfont \quad\apbbox{L} \\ \hline

		Convolution & 12 & 38.6 & 59.7 & 42.0 & 23.6 & 42.3 & 50.0 \\ 
		PSSA & 12 & 38.7\cgap{+}{0.1} & 60.4\cgap{+}{0.7} & 42.2\cgap{+}{0.2} & 23.8\cgap{+}{0.2} & 42.3\cgap{+}{0.0} & 49.1\cgap{-}{0.9} \\ 
	    X-volution & 12 & 40.1\cgap{+}{1.5} & 61.3\cgap{+}{1.6} & 43.6\cgap{+}{1.6} & 24.0\cgap{+}{0.4} & 43.6\cgap{+}{1.3} & 51.4\cgap{+}{1.4} \\
	
		\cline{1-8}
		
		Convolution & 24 & 40.1 & 61.1 & 43.7 & 23.9 & 43.7 & 52.2 \\ 
		PSSA & 24 & 40.4\cgap{+}{0.3} & 61.9\cgap{+}{0.8} & 44.1\cgap{+}{0.4} & 24.9\cgap{+}{1.0} & 44.1\cgap{+}{0.4} & 51.4\cgap{-}{0.8} \\ 
	    X-volution  & 24 & 42.0\cgap{+}{1.9}  & 63.8\cgap{+}{2.7} & 45.9\cgap{+}{2.2} & 26.4\cgap{+}{2.5} & 45.5\cgap{+}{1.8} & 54.2\cgap{+}{2.0} \\ 
	    \cline{1-8}
	    
		Convolution & 36 & 40.7 & 61.5 & 44.5 & 24.5 & 44.5 & 52.5 \\ 
		PSSA & 36 & 41.2\cgap{+}{0.5} & 62.8\cgap{+}{1.3} & 44.9\cgap{+}{0.4} & 25.4\cgap{+}{0.9} & 44.9\cgap{+}{0.4} & 52.3\cgap{-}{0.2} \\  
	    X-volution & 36 & 42.7\cgap{+}{2.0} & 63.9\cgap{+}{2.4} & 46.3\cgap{+}{1.8} & 26.5\cgap{+}{2.0} & \textbf{46.3}\cgap{+}{1.8} & 54.8\cgap{+}{2.3} \\
	    \cline{1-8}
	    
		Convolution & 72 & 41.0   & 61.7 & 44.8 & 24.2 & 44.3 & 53.1 \\ 
		PSSA & 72 & 41.7\cgap{+}{0.7} & 63.1\cgap{+}{1.4} & 45.4\cgap{+}{0.6} & 25.8\cgap{+}{1.6} & 45.1\cgap{+}{0.8} & 53.4\cgap{+}{0.3} \\ 
	    X-volution & 72 & \textbf{42.8}\cgap{+}{1.8} & \textbf{64.0}\cgap{+}{2.3} & \textbf{46.4}\cgap{+}{1.6} & \textbf{26.9}\cgap{+}{2.7} & 46.0\cgap{+}{1.7}   & \textbf{55.0}\cgap{+}{1.9}   \\
		\Xhline{1.0pt}
		
\end{tabular}
\end{threeparttable}
\end{spacing}
\end{table*}

\begin{table*}[!htb]\centering
 \caption{Learning rate schedules for the 1x, 2x, 3x and 6x settings.}
 \label{tab:lr schedule}
 \begin{spacing}{1.1}
 \small
 \setlength\tabcolsep{3pt}
 \begin{tabular}{c|c|c|c}
        \Xhline{1.0pt}
        Setting
        & Epochs
        & Training Steps 
        & Learning Rate Schedule \\ \hline

		1x & 12 & 90k & [60k,\; 80k]\\
		2x & 24 & 180k & [120,\; 160k]\\
		3x & 36 & 270k & [210k,\; 250k]\\
		6x & 72 & 540k & [420k,\; 500k]\\
		\Xhline{1.0pt}
\end{tabular}
\vspace{-1mm}
\end{spacing}
\end{table*}

Tab.~\ref{tab:more det results} compares the results of different operators under different training schedules. Convolution denotes the ResNet-50 baseline. We can clearly observe that our X-volution brings steady and significant improvements over the convolution, suggesting that the combination of convolution and self-attention contributes to better visual understanding. Under \textbf{1x} setting, PSSA achieves comparable performance comparing to convolution, and the performance of PSSA is obviously better than that of convolution as the training steps increase. Moreover, the gain of X-volution also increases with more training epochs, and the most significant gain (\emph{i.e.}, +2.0 AP) is obtained under \textbf{3x} setting. This phenomenon indicates that self-attention operator needs a longer training circle to achieve better performance than convolution, and the multi-branch topology allows the X-volution module to possess well-conditioned optimization properties from convolution and also benefit from long-range interactions introduced by self-attention, thus showing stronger learning ability.

\bibliographystyle{plain}
\bibliography{main}

\end{document}